\documentclass[sigconf]{acmart}

\usepackage{amsmath}
\usepackage{algorithm}
\usepackage{algorithmic}
\usepackage{amsfonts}
\usepackage{bbding}
\usepackage{bm}
\usepackage{multirow}
\usepackage{subfigure}
\usepackage{color}
\usepackage{balance}

\definecolor{darkgreen}{rgb}{0.0, 0.5, 0.0}
\newcommand\true{\textcolor{darkgreen}{\textnormal{\Checkmark}}}
\definecolor{darkred}{rgb}{0.76, 0.13, 0.28}
\newcommand\false{\textcolor{darkred}{\textnormal{\XSolidBrush}}}

\AtBeginDocument{%
  }

\copyrightyear{2026}
\acmYear{2026}
\setcopyright{cc}
\setcctype{by}
\acmConference[KDD '26]{Proceedings of the 32nd ACM SIGKDD Conference on Knowledge Discovery and Data Mining V.2}{August 09--13, 2026}{Jeju Island, Republic of Korea}
\acmBooktitle{Proceedings of the 32nd ACM SIGKDD Conference on Knowledge Discovery and Data Mining V.2 (KDD '26), August 09--13, 2026, Jeju Island, Republic of Korea}
\acmDOI{10.1145/3770855.3818841}
\acmISBN{979-8-4007-2259-2/2026/08}

\begin{document}

\title[EnergyMamba]{EnergyMamba: An Uncertainty-Aware Graph-Enhanced Selective State Space Model for Energy Consumption Prediction}

\author{Dahai Yu}
\affiliation{
  \institution{Florida State University}
  \city{Tallahassee, Florida}
  \country{USA}
}
\email{dahai.yu@fsu.edu}

\author{Rongchao Xu}
\affiliation{
  \institution{Florida State University}
  \city{Tallahassee, Florida}
  \country{USA}
}
\email{rx21a@fsu.edu}

\author{Lin Jiang}
\affiliation{
  \institution{Florida State University}
  \city{Tallahassee, Florida}
  \country{USA}
}
\email{lin.jiang@fsu.edu}

\author{Guang Wang}
\authornote{Prof. Guang Wang is the corresponding author.}
\affiliation{
  \institution{Florida State University}
  \city{Tallahassee, Florida}
  \country{USA}
}
\email{guang@cs.fsu.edu}

\renewcommand{\shortauthors}{Yu et al.}

\begin{abstract}
Energy consumption prediction is essential for efficient grid management, demand-side optimization, and sustainable energy planning. 
Although advanced machine learning methods have been employed for better prediction performance, existing works have two key limitations: (1) they usually formulate this task as a purely time-series prediction problem without explicitly modeling the spatial dependencies among different regions, and (2) they fail to provide reliable predictions with uncertainty estimates under abnormal situations such as extreme weather events.
To advance existing research, we propose EnergyMamba, an uncertainty-aware spatiotemporal learning framework for accurate and reliable energy consumption prediction, which comprises two key components: (i) a novel Graph-Enhanced Selective State Space Model (GE-Mamba) that injects spatial context learned from the grid topology into the temporal dynamics, enabling coupled spatiotemporal modeling, and (ii) an Adaptive Sequential Conformalized Quantile Regression (AS-CQR) module, which includes locally adaptive normalization and an online feedback mechanism to dynamically calibrate prediction intervals under potential distribution shifts.
We evaluate EnergyMamba on four large-scale real-world datasets from Florida, New York, and California. 
Results show EnergyMamba achieves around 5\% improvement in prediction accuracy and 6\% improvement in uncertainty quantification over 15 state-of-the-art baselines.
\end{abstract}



\maketitle

\section{Introduction}\label{introduction}
Energy consumption prediction has attracted substantial attention from both industry and academia due to its critical role in supporting a wide range of practical applications with significant societal impacts. Accurate and reliable energy consumption prediction is essential for efficient grid management~\cite{JHA2021107479}, demand-side optimization~\cite{binbusayyis2025energy}, and sustainable energy planning~\cite{peteleaza2024electricity}. 
It also contributes to emergency preparedness~\cite{zhong2010challenges} by helping stakeholders anticipate potential surges in energy consumption during extreme weather events, ensuring the stability and resilience of the power grid. 

Driven by its practical significance, energy consumption prediction has been studied using a variety of approaches, including traditional statistical models~\cite{8666125,8766698}, classical machine learning methods~\cite{deng2026adaptive,lim2021temporal,li2022spatiotemporal,piccialli2020deep}, general deep learning methods~\cite{bouktif2020multi,yu2025uqgnn,yu2026trustenergy}, Transformers~\cite{alexandrov2020gluonts,nie2022time}, and Diffusion~\cite{jiang2025hcride,jiang2025uncertainty}. More recently, Foundation Models~\cite{tu2024powerpm} and large language models (LLMs)~\cite{10446624,zhang2026finsentllm,liang2025energygpt} have been explored to model complex energy systems.
However, there are two key limitations of existing work.
First, most existing work formulates energy consumption prediction as a purely time-series forecasting problem, neglecting the intrinsic spatial dependencies among different regions or grid zones that could provide critical contextual information for accurate prediction. 
Second, state-of-the-art approaches fail to continuously provide reliable predictions with uncertainty estimates, especially under abnormal situations such as extreme weather events, leading to overconfident and potentially misleading predictions.

In this work, we aim to develop an uncertainty-aware spatiotemporal framework for energy consumption prediction that explicitly models spatial dependencies while providing reliable uncertainty estimates. Nevertheless, there are two key challenges to achieving this.
First, energy consumption patterns exhibit strong correlations with geographical factors, temporal factors, and building characteristics (e.g., commercial vs. residential). These characteristics lead to highly heterogeneous consumption patterns across regions and time periods, making it difficult for existing models to effectively capture fine-grained spatiotemporal dependencies.
Second, energy consumption behaviors and demand distributions may shift dramatically during extreme events, which makes it challenging to continuously provide reliable predictions and uncertainty estimates.

To address the above challenges and advance existing research, we propose EnergyMamba, an uncertainty-aware spatiotemporal learning framework for accurate and reliable
energy consumption prediction. 
First, to address the spatiotemporal heterogeneity challenge, in EnergyMamba, we design \textbf{Graph-Enhanced Selective
State Space Model (GE-Mamba)}, a novel architecture that injects spatial context learned from grid topology into a bidirectional Mamba, organized within a U-Net structure. GE-Mamba leverages Graph Convolutional Networks to extract spatial context from the grid topology and injects this context directly into an efficient Selective State Space Model for temporal dynamics modeling. 
This design is motivated by Kirchhoff's circuit laws: consumption changes at one node propagate to connected nodes through power flow equations, so conditioning temporal dynamics on spatial context enables more accurate modeling. 
Second, to address the uncertainty quantification challenge, we propose \textbf{Adaptive Sequential Conformalized Quantile Regression (AS-CQR)}. Unlike standard conformal prediction that assumes data exchangeability, AS-CQR is designed for non-stationary time series. It includes (i) a locally adaptive nonconformity measure that normalizes residuals by the predicted interval width, ensuring scale-invariant calibration across different regions, and (ii) an online feedback mechanism that dynamically adjusts the target quantile level based on recent coverage performance, enabling rapid adaptation to distribution shifts. 

The key contributions of this paper are as follows: 
\begin{itemize}
    \item \textbf{Conceptually}, different from existing works that treat energy consumption prediction as a purely time-series prediction problem, we formulate it as an uncertainty-aware spatiotemporal prediction task with explicit spatial dependency modeling and reliable uncertainty quantification.
    
    \item \textbf{Technically}, we propose EnergyMamba, an uncertainty-aware spatiotemporal learning framework comprising: (i) GE-Mamba, a novel architecture that injects spatial context learned from grid topology into a bidirectional Mamba, organized within a U-Net structure; and (ii) AS-CQR, a distribution-free uncertainty quantification method with locally adaptive normalization and online feedback calibration for reliable prediction under non-stationary conditions.
    
    \item \textbf{Empirically}, we evaluate EnergyMamba on four real-world datasets from Florida, New York, and California by comparing it with 15 state-of-the-art baselines across five metrics. Extensive results demonstrate that EnergyMamba achieves around 5\% higher prediction accuracy and 6\% better uncertainty quantification than the best baseline. Our implementation is available at \url{https://github.com/UFOdestiny/EnergyMamba}.
\end{itemize}

\section{Data Analysis and Motivation }\label{sec:data}
In this section, we conduct a data-driven analysis to highlight key findings that motivate our design. 
More detailed descriptions of data collection, preprocessing, and management procedures are described in Appendix \ref{sec:app_data}.
\subsection{Data Description}
In this project, we are collaborating with a municipal utility provider in Florida.
We have access to the utility data collected from over 60,000 smart meters in Leon County, FL. It provides us with household-level energy consumption time series at a 30-minute interval. The raw data are aggregated at the Census Block Group (CBG) level to prevent the disclosure of personally identifiable information and safeguard user privacy.

\subsection{Data-driven Insights}

\subsubsection{Spatial autocorrelation and local heterogeneity.}
\begin{figure}[ht]
\centering
\subfigure[Moran's~I scatter plot.]{\label{fig:moran_scatter}
\includegraphics[width=0.48\linewidth]{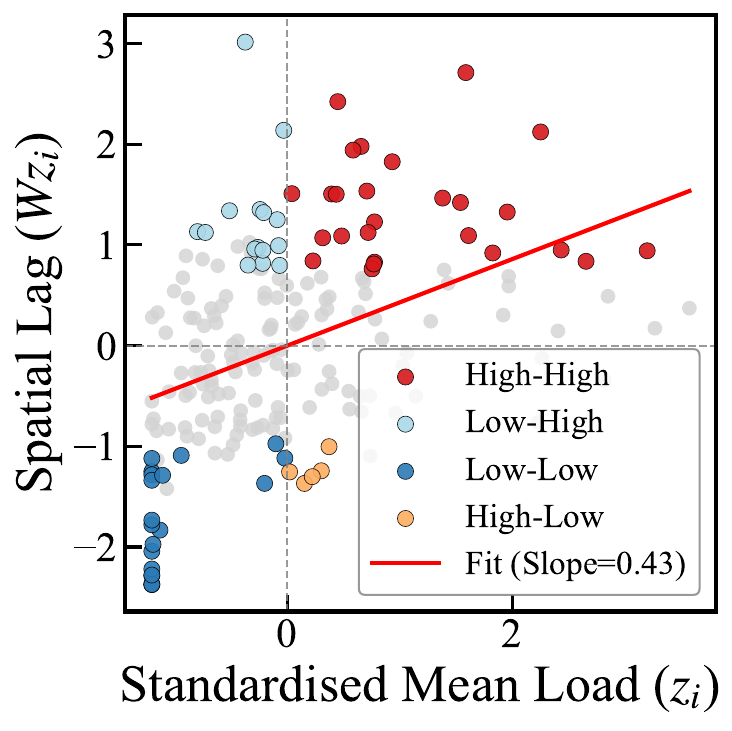}}
\subfigure[LISA cluster map.]{\label{fig:lisa_map}
\includegraphics[width=0.48\linewidth]{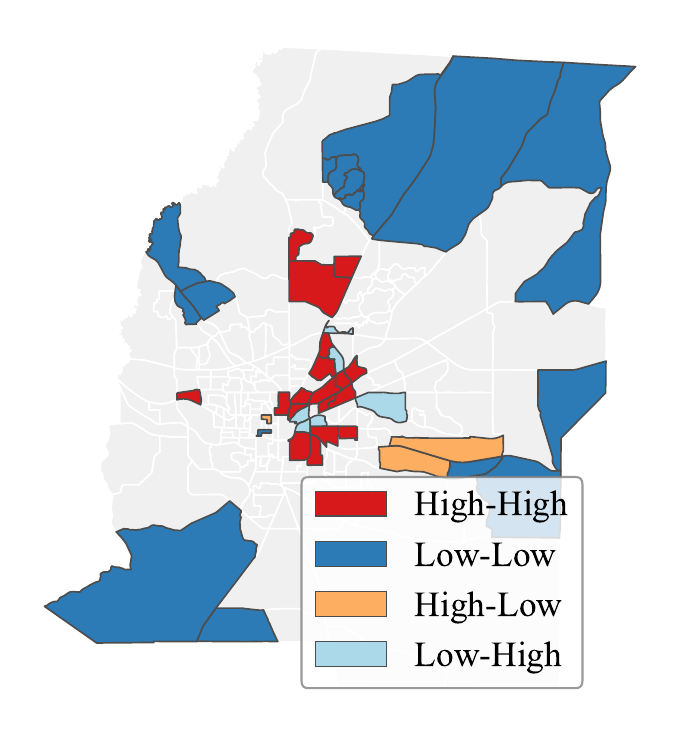}}
\caption{Spatial autocorrelation. \textbf{High-Low} means High-consumption CBG surrounded by Low-usage neighbors.}
\Description{Two panels summarize spatial autocorrelation in the Florida data: a Moran's I scatter plot on the left and a LISA cluster map on the right.}
\label{fig:spatial_autocorrelation}
\end{figure}

We employ Moran's I statistics~\cite{moran1950notes} to perform both global and local spatial autocorrelation analyses of energy consumption at the CBG level. 
Figure~\ref{fig:moran_scatter} visualizes the relationship between a CBG’s own state and that of its neighboring CBGs, in which $z_i$ represents the standardized mean load of the $i$-th CBG, while spatial lag ($Wz_i$) denotes the weighted average load of its neighboring CBGs computed using a row-normalized spatial weight matrix $W$.
The plot divides the data into four quadrants reflecting distinct local spatial correlations:
(1) \textbf{High-High} and \textbf{Low-Low} indicating spatial clusters where a CBG shares similar patterns with its neighbors (positive correlation); and
(2) \textbf{High-Low} and \textbf{Low-High} highlighting spatial dispersion where a CBG deviates from its surroundings (negative correlation).
The fitted slope of the Moran's I plot in Figure~\ref{fig:moran_scatter} is 0.429 ($p < 0.001$), indicating statistically significant positive spatial autocorrelation. From a physical systems perspective, this spatial dependence is not incidental but arises from fundamental circuit constraints.
Specifically, Kirchhoff's Current Law enforces nodal power balance, implying that a local load perturbation at a CBG must be redistributed through adjacent links. 
This redistribution propagates along the network topology, inducing correlated load deviations at electrically connected neighbors.

Beyond global correlation, Figure~\ref{fig:lisa_map} projects these statistical associations onto the physical regions to reveal local heterogeneity.
\textbf{(High-High/Low-Low):} High-High clusters align with the dense urban core, reflecting potential electrical coupling, while Low-Low clusters dominate the rural periphery with sparse connectivity.
\textbf{(High-Low/Low-High):} The map also identifies spatial regions that defy the global trend. For instance, a High-Low node likely identifies a CBG containing concentrated commercial or industrial loads (e.g., a shopping complex or a hospital district) surrounded by lower-density residential neighborhoods.
These topological mismatches challenge purely time-series models but provide meaningful signals for spatially aware models to distinguish between normal fluctuations and specific load types.

\begin{figure}[ht]
\centering
\subfigure[Load--residual scaling.]{
\label{fig:hetero_scatter}
\includegraphics[width=0.48\linewidth]{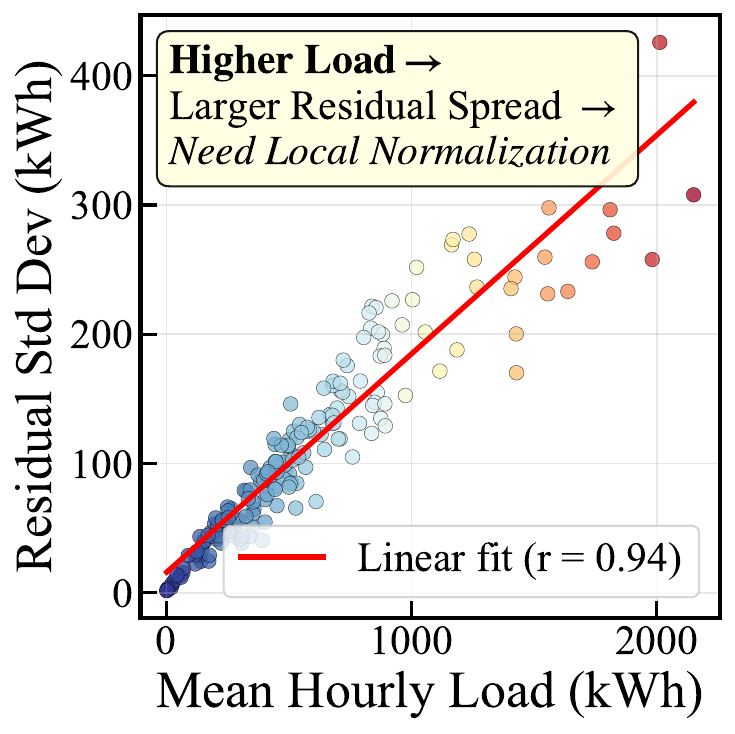}}
\subfigure[Locally adaptive normalization.]{
\label{fig:normalize_flat}
\includegraphics[width=0.48\linewidth]{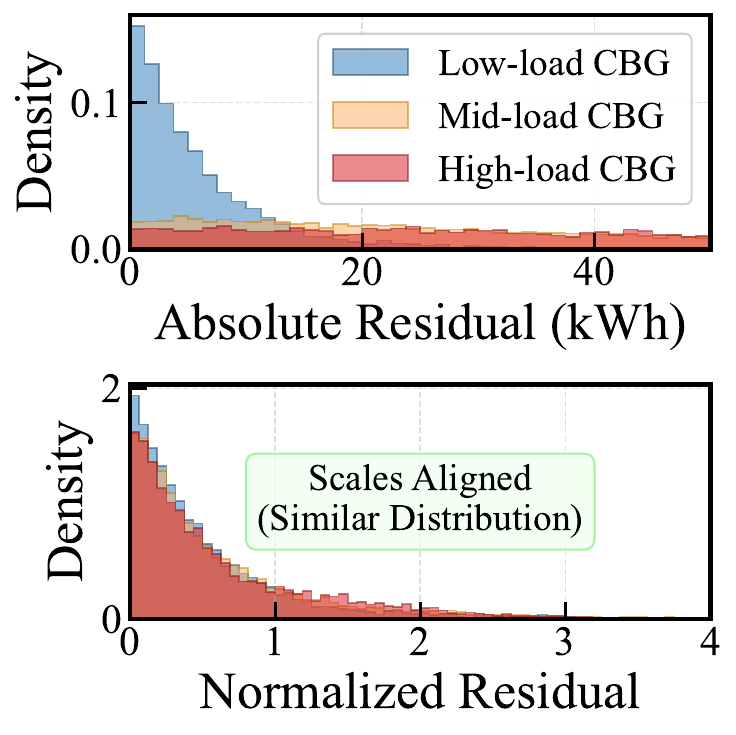}}
\caption{Uncertainty in energy consumption prediction.}
\Description{Two panels illustrate uncertainty behavior: load-residual scaling on the left and the effect of locally adaptive normalization on the right.}
\label{fig:uncertainty_analysis}
\end{figure}

\subsubsection{Heteroscedastic uncertainty scales with load magnitude.}
Figure~\ref{fig:hetero_scatter} reveals a strong linear relationship ($r = 0.94$) between the mean hourly load and the standard deviation of prediction residuals, which indicates heteroscedasticity, i.e., the variability of electricity consumption is not constant but proportional to the magnitude of usage; regions with higher base loads naturally exhibit larger absolute fluctuations and prediction errors.
As illustrated in Figure~\ref{fig:normalize_flat}, this scale dependence causes the raw absolute residual distributions to differ significantly across load tiers.
By normalizing these residuals by their respective local standard deviations, the distributions across low, mid, and high-load groups collapse onto a nearly identical curve.
This transformation stabilizes the variance, effectively converting the heteroscedastic errors into a scale-independent distribution suitable for standardized analysis.

These insights highlight two essential requirements for energy consumption modeling and prediction. First, strong \textbf{spatial dependencies} imply that prediction should account for interactions across neighboring regions. 
Second, the observed heteroscedasticity and non-stationarity indicate that uncertainty estimation must adapt to \textbf{changing load levels and distribution shifts}.
\section{Problem Formulation}\label{sec:problem-formulation}

\subsection{Graph Construction}\label{subsec:graph-construction}
We model the energy systems as a graph $\mathcal{G}=(\mathbf{V},\mathbf{E},\mathbf{A})$, where $\mathbf{V}=\{v_1,\ldots,v_N\}$ denotes $N$ nodes (e.g., Census Block Groups or counties), $\mathbf{E}$ is the edge set constructed via spatial proximity to approximate physical coupling, and $\mathbf{A}\in\mathbb{R}^{N\times N}$ is the weighted adjacency matrix. 
We construct $\mathbf{A}$ using a Gaussian kernel with sparsity thresholding:
\begin{equation}
\mathbf{A}_{ij}=
\begin{cases}
\exp\!\left(-d_{ij}^{2}/\sigma^{2}\right), & i\neq j~\text{and}~\exp\!\left(-d_{ij}^{2}/\sigma^{2}\right)\ge \epsilon,\\[4pt]
0, & \text{otherwise},
\end{cases}
\label{eq:adjacency}
\end{equation}
where $d_{ij}$ is the centroid distance between nodes $v_i$ and $v_j$, $\sigma>0$ controls spatial decay, and $\epsilon\in[0,1]$ is a sparsity threshold that removes weak connections.

\subsection{Energy Consumption Prediction}\label{subsec:prediction}
Let $\mathbf{X}\in\mathbb{R}^{N\times T}$ denote the input matrix containing energy consumption readings of all $N$ nodes over $T$ historical time steps. Given the historical observations $\mathbf{X}$ and the adjacency matrix $\mathbf{A}$, the goal is to predict future consumption for the next $T_{\text{out}}$ time steps.

\subsubsection{Point Prediction}\label{subsubsec:deterministic}
Point prediction learns a mapping $f$ from inputs to deterministic predictions:
\begin{equation}
\{\mathbf{X}, \mathbf{A}\}\xrightarrow{~f~}\hat{\mathbf{Y}}\in\mathbb{R}^{N\times T_{\text{out}}}.
\label{eq:deterministic}
\end{equation}

\subsubsection{Probabilistic Prediction with Uncertainty Quantification}\label{subsubsec:probabilistic}
In this work, we focus on probabilistic prediction, which quantifies uncertainty using prediction intervals. Given a target miscoverage rate $\alpha$ (e.g., $\alpha=0.1$ for 90\% coverage), we learn a mapping function $\mathcal{F}$ to predict lower, upper, and median quantiles:
\begin{equation}
\{\mathbf{X}, \mathbf{A}\}\xrightarrow{~\mathcal{F}~}
\big[\hat{q}_{\text{lo}},\,\hat{q}_{\text{up}},\,\hat{q}_{\text{mi}}\big] \in \mathbb{R}^{N\times T_{\text{out}}},
\label{eq:probabilistic}
\end{equation}
where $\hat{q}_{\text{lo}} = \hat{q}_{\alpha/2}(\mathbf{X})$, $\hat{q}_{\text{up}} = \hat{q}_{1-\alpha/2}(\mathbf{X})$, and $\hat{q}_{\text{mi}} = \hat{q}_{0.5}(\mathbf{X})$ represent the lower, upper bounds and median of the prediction interval, respectively. The prediction interval $\hat{C}(\mathbf{X}) = [\hat{q}_{\text{lo}}, \hat{q}_{\text{up}}]$ should satisfy the coverage guarantee:
\begin{equation}
\mathbb{P}\left(Y \in \hat{C}(\mathbf{X})\right) \geq 1 - \alpha,
\label{eq:coverage}
\end{equation}
where $Y \in \mathbb{R}$ is the ground truth value for an individual prediction target. The median $\hat{q}_{\text{mi}}$ serves as the point prediction $\hat{\mathbf{Y}}$.

\begin{figure*}[htpb]
    \centering
\includegraphics[width=\linewidth]{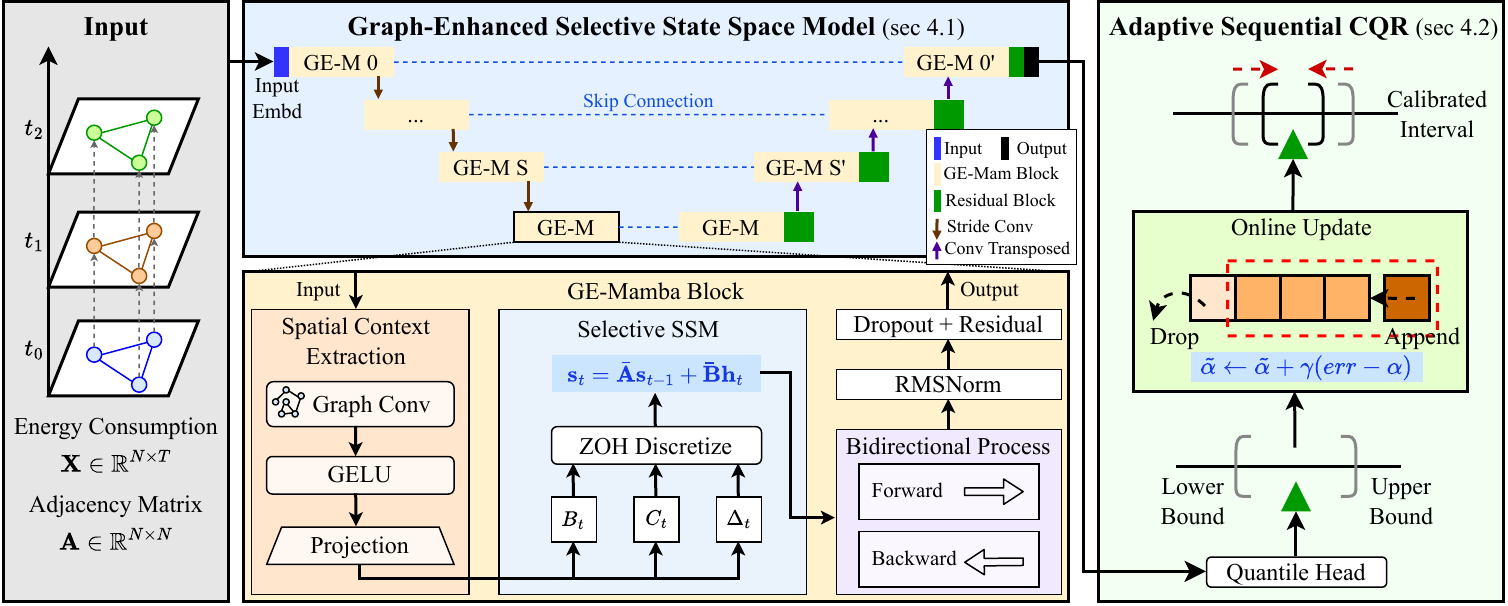}
  \caption{Framework of EnergyMamba, which consists of two key components: (i) \textbf{Graph-Enhanced Selective State Space Model (GE-Mamba)}, and (ii) \textbf{Adaptive Sequential Conformalized Quantile Regression (AS-CQR)}.}
  \Description{Overview of the EnergyMamba framework, showing the GE-Mamba backbone and the AS-CQR calibration module and how they connect from input sequences to point and interval predictions.}
  \label{fig:framework}
\end{figure*}

\section{Methodology}
\label{sec:methodology}
The empirical insights derived in Section~\ref{sec:data} directly inform the design of EnergyMamba, which aims to bridge data-driven learning with physical system principles. 
First, the spatial dependencies governed by Kirchhoff's laws motivate the (i) \textbf{Graph-Enhanced Selective State Space Model (GE-Mamba)}. Instead of treating spatial aggregation as generic feature extraction, we design the GCN module as a learnable neural surrogate for power flow equations, conditioning Mamba's state transitions on physics-informed neighborhood context to simulate how local load perturbations propagate through the grid topology. Second, the load-dependent heteroscedasticity and non-stationarity motivate (ii) \textbf{Adaptive Sequential Conformalized Quantile Regression (AS-CQR)}. Recognizing that energy systems exhibit both aleatoric uncertainty (scaling with load magnitude due to physical losses) and epistemic uncertainty (arising from distribution shifts during extreme events), AS-CQR employs scale-invariant normalization to handle the former and an online feedback loop to dynamically adapt to the latter.

\subsection{Graph-Enhanced Selective State Space Model}
\label{subsec:gemamba}
In energy systems, consumption changes at one substation propagate to connected nodes through power flow equations; ignoring this coupling leads to suboptimal temporal modeling. We address this through a novel Graph-Enhanced Mamba architecture that injects spatial context into the selective state space model (SSM), organized within a U-Net encoder-decoder structure that enables multi-scale temporal feature extraction with skip connections for preserving fine-grained patterns. 

\subsubsection{Input Embedding}
\label{subsubsec:embedding}
The input matrix $\mathbf{X} \in \mathbb{R}^{N \times T}$ contains energy consumption readings of $N$ nodes over $T$ time steps. Before feeding into the U-Net architecture, each node's time series is projected into a latent space of dimension $D$. Specifically, each value $x_{n,t} \in \mathbb{R}$ is independently projected to a $D$-dimensional vector via a shared linear layer:
\begin{equation}
\mathbf{h}^{(0)}_{n,t} = \mathbf{W}_{\text{emb}} x_{n,t} + \mathbf{e}_{\text{pos},t},
\label{eq:input-embedding}
\end{equation}
where $\mathbf{W}_{\text{emb}} \in \mathbb{R}^{D \times 1}$ is a learnable projection that maps each scalar to a $D$-dimensional embedding, $\mathbf{e}_{\text{pos},t} \in \mathbb{R}^{D}$ is the learnable temporal position embedding at time step $t$, $D$ is the hidden dimension, and stacking all $\mathbf{h}^{(0)}_{n,t}$ yields $\mathbf{H}^{(0)} \in \mathbb{R}^{N \times T \times D}$ as the initial hidden representation for the first encoder stage.

\subsubsection{U-Net Architecture with GE-Mamba Backbone}
\label{subsubsec:unet}

To capture multi-scale temporal patterns, including short-term fluctuations and long-term trends, we organize GE-Mamba blocks within a U-Net style encoder-decoder architecture.
The model comprises $S$ encoder stages, a bottleneck, and $S$ decoder stages. Each encoder stage stacks $K$ GE-Mamba blocks followed by temporal downsampling, which halves the temporal resolution and doubles the hidden dimension to $D^{(s)} = 2^{s-1}D$. The bottleneck operates on the most compressed representation, while the decoder symmetrically upsamples features using transposed convolutions and skip connections. These skip connections fuse coarse temporal patterns (e.g., weekly trends) with fine-grained variations (e.g., hourly peaks), preserving abnormal events such as sudden demand spikes.

\subsubsection{GE-Mamba Block}
\label{subsubsec:gemamba-block}

Each GE-Mamba block integrates spatial context extraction and temporal modeling with residual connections. The block takes hidden representations $\mathbf{H}^{(l-1)}$ from the previous layer and outputs updated representations $\mathbf{H}^{(l)}$. 

\textbf{Spatial Context Extraction.}
To extract spatial context from the grid topology, we employ a Graph Convolutional Network (GCN) that aggregates information from neighboring nodes. This design is motivated by Kirchhoff's laws in electrical networks: the power balance at each node depends on flows from all connected edges \cite{quintela2009general}.
For each time step $t$, let $\mathbf{H}_t \in \mathbb{R}^{N \times D}$ denote the hidden representation. The spatial context $\mathbf{Z}_t \in \mathbb{R}^{N \times D}$ is computed as:
\begin{equation}
\mathbf{Z}_t = \text{GCN}(\mathbf{H}_t, \mathbf{A}) = \text{GELU}\left(\tilde{\mathbf{D}}^{-\frac{1}{2}} \tilde{\mathbf{A}} \tilde{\mathbf{D}}^{-\frac{1}{2}} \mathbf{H}_t \mathbf{W}_{\text{gcn}}\right),
\label{eq:gcn}
\end{equation}
where $\tilde{\mathbf{A}} = \mathbf{A} + \mathbf{I}_N$ is the adjacency matrix augmented with self-loops, $\tilde{\mathbf{D}} \in \mathbb{R}^{N \times N}$ is the diagonal degree matrix with $\tilde{\mathbf{D}}_{ii} = \sum_j \tilde{\mathbf{A}}_{ij}$, $\mathbf{W}_{\text{gcn}} \in \mathbb{R}^{D \times D}$ is a learnable weight matrix, and $\text{GELU}(\cdot)$ is the Gaussian Error Linear Unit activation function.
The GCN is applied independently at each time step, producing a spatial context sequence $\mathbf{Z} = [\mathbf{Z}_1, \ldots, \mathbf{Z}_T] \in \mathbb{R}^{N \times T \times D}$ that is then fed into the SSM along with $\mathbf{H}^{(l-1)}$ for joint spatiotemporal modeling.
The symmetric normalization $\tilde{\mathbf{D}}^{-\frac{1}{2}} \tilde{\mathbf{A}} \tilde{\mathbf{D}}^{-\frac{1}{2}}$ ensures numerically stable message passing across nodes with varying degrees, which is important for power systems where hub substations connect to many downstream nodes while peripheral nodes have fewer connections.

\textbf{Graph-Enhanced Selective State Space Model.}
The Selective State Space Model (Mamba)~\cite{gu2024mamba} provides an efficient mechanism for temporal sequence modeling with linear complexity $\mathcal{O}(T)$. The key innovation of our approach is to condition the SSM dynamics on spatial context, enabling the model to adapt its temporal processing based on the state of neighboring nodes.
The SSM operates independently on each of the $D$ hidden channels. For a single channel, the continuous-time dynamics are defined by a latent state $\mathbf{s}(t) \in \mathbb{R}^{D_s}$, where $D_s$ is the state dimension:
\begin{align}
\mathbf{s}'(t) &= \mathbf{A}_{\text{ssm}}\mathbf{s}(t) + \mathbf{B}x(t), \quad y(t) = \mathbf{C}\mathbf{s}(t),\label{eq:ssm-continuous-state}
\end{align}
where $\mathbf{A}_{\text{ssm}} \in \mathbb{R}^{D_s \times D_s}$ is the state transition matrix, $\mathbf{B} \in \mathbb{R}^{D_s \times 1}$ is the input projection matrix, $\mathbf{C} \in \mathbb{R}^{1 \times D_s}$ is the output projection matrix, and $x(t) \in \mathbb{R}$ is the input for that channel.
Since the continuous-time formulation in Eq.~\eqref{eq:ssm-continuous-state} involves derivatives that are not directly applicable to sampled observations such as hourly energy readings, we apply zero-order hold (ZOH) discretization with step size $\Delta > 0$ to obtain a recurrence relation suitable for discrete time-series data:
\begin{align}
\bar{\mathbf{A}} &= \exp(\Delta \mathbf{A}_{\text{ssm}}), \quad
\bar{\mathbf{B}} = (\Delta \mathbf{A}_{\text{ssm}})^{-1}(\exp(\Delta \mathbf{A}_{\text{ssm}}) - \mathbf{I}) \cdot \Delta \mathbf{B},
\end{align}
where $\bar{\mathbf{A}}$ and $\bar{\mathbf{B}}$ are the discretized state transition and input matrices. Aggregating over all $D$ channels, the discrete recurrence for the full hidden state is:
\begin{align}
\mathbf{s}_t &= \bar{\mathbf{A}} \mathbf{s}_{t-1} + \bar{\mathbf{B}} \mathbf{h}_t, \quad
\tilde{\mathbf{h}}_t = \mathbf{C} \mathbf{s}_t,
\end{align}
where $\mathbf{h}_t \in \mathbb{R}^{D}$ is the input hidden state at time $t$ and $\tilde{\mathbf{h}}_t \in \mathbb{R}^{D}$ is the output.

\textbf{Spatial-Conditioned Selectivity.}
Building upon the Mamba structure's input-dependent nature, we enhance its selective information propagation by introducing a mechanism that conditions on both temporal inputs and spatial contexts. Specifically, the static $\mathbf{B}$, $\mathbf{C}$, and $\Delta$ are replaced by time-varying, input-dependent counterparts:
\begin{align}
\mathbf{B}_t &= \mathbf{W}_B[\mathbf{h}_t \| \mathbf{z}_t] + \mathbf{b}_B, \label{eq:selective-B}\\
\mathbf{C}_t &= \mathbf{W}_C[\mathbf{h}_t \| \mathbf{z}_t] + \mathbf{b}_C, \label{eq:selective-C}\\
\Delta_t &= \text{Softplus}(\mathbf{W}_\Delta[\mathbf{h}_t \| \mathbf{z}_t] + b_\Delta), \label{eq:selective-delta}
\end{align}
where $[\cdot \| \cdot]$ denotes concatenation, $\mathbf{z}_t \in \mathbb{R}^{D}$ is the spatial context from Eq.~\eqref{eq:gcn}, $\mathbf{W}_B, \mathbf{W}_C \in \mathbb{R}^{D_s \times 2D}$ and $\mathbf{W}_\Delta \in \mathbb{R}^{1 \times 2D}$ are learnable weight matrices, $\mathbf{b}_B, \mathbf{b}_C \in \mathbb{R}^{D_s}$ and $b_\Delta \in \mathbb{R}$ are learnable biases, and $\text{Softplus}(x) = \log(1 + e^x)$ ensures positivity of the step size. This spatial conditioning has a physical interpretation: the step size $\Delta_t$ controls how quickly the model forgets past states, and by conditioning on spatial context, nodes experiencing similar patterns in their neighborhood exhibit coordinated memory behavior.

\textbf{Bidirectional Processing (BIP).}
Energy time series exhibit both causal dependencies (past events affect future consumption) and contextual dependencies (understanding patterns requires seeing the full sequence). We employ bidirectional processing:
\begin{equation}
\text{BIP}(\mathbf{H}, \mathbf{Z}) = \text{Mamba}_{\rightarrow}(\mathbf{H}, \mathbf{Z}) + \text{Mamba}_{\leftarrow}(\text{Flip}(\mathbf{H}), \text{Flip}(\mathbf{Z})),
\label{eq:bidirectional-mamba}
\end{equation}
where $\text{Mamba}_{\rightarrow}$ processes the sequence forward in time, $\text{Mamba}_{\leftarrow}$ processes backward, and $\text{Flip}(\cdot)$ reverses the temporal dimension.

\textbf{Block Integration.}
For a single block at layer $l$, we have
\begin{align}
\mathbf{Z}^{(l)} &= \text{GCN}(\text{RMSNorm}(\mathbf{H}^{(l-1)}), \mathbf{A}), \label{eq:block-gcn}\\
\mathbf{H}^{(l)} &= \mathbf{H}^{(l-1)} + \text{Dropout}(\text{BIP}(\text{RMSNorm}(\mathbf{H}^{(l-1)}), \mathbf{Z}^{(l)})), \label{eq:block-mamba}
\end{align}
where $\text{RMSNorm}(\cdot)$ is Root Mean Square Layer Normalization~\cite{zhang2019root}.

\subsubsection{Output Projection}
\label{subsubsec:output}

The final decoder output $\mathbf{H}^{(1)}_{\text{dec}} \in \mathbb{R}^{N \times T \times D}$ is projected to produce both point predictions and quantile estimates, where $D$ is the hidden dimension. We extract the last-time-step representation, denoted by $\mathbf{H}_{\text{out}} \in \mathbb{R}^{N \times D}$, and apply three separate linear projections:
\begin{align}
\hat{\mathbf{Y}} &= \text{RMSNorm}(\mathbf{H}_{\text{out}}) \mathbf{W}_{\text{mi}} + \mathbf{b}_{\text{mi}}, \label{eq:output-mean}\\
\hat{q}_{\text{lo}} &= \text{RMSNorm}(\mathbf{H}_{\text{out}}) \mathbf{W}_{\text{lo}} + \mathbf{b}_{\text{lo}}, \label{eq:output-lo}\\
\hat{q}_{\text{up}} &= \text{RMSNorm}(\mathbf{H}_{\text{out}}) \mathbf{W}_{\text{up}} + \mathbf{b}_{\text{up}}, \label{eq:output-up}
\end{align}
where $\mathbf{W}_{\text{mi}}, \mathbf{W}_{\text{lo}}, \mathbf{W}_{\text{up}} \in \mathbb{R}^{D \times T_{\text{out}}}$ are learnable weight matrices, $\mathbf{b}_{\text{mi}}, \mathbf{b}_{\text{lo}}, \mathbf{b}_{\text{up}} \in \mathbb{R}^{T_{\text{out}}}$ are learnable biases, and $T_{\text{out}}$ is the prediction horizon.

\subsubsection{Training Objective}
\label{subsubsec:loss}

GE-Mamba is trained end-to-end using a composite quantile regression loss. For each quantile level $\tau \in \{\alpha/2,\, 0.5,\, 1-\alpha/2\}$, we adopt the pinball loss:
\begin{equation}
\mathcal{L}_{\tau}(y, \hat{q}_\tau) =
\begin{cases}
\tau \cdot (y - \hat{q}_\tau), & y \geq \hat{q}_\tau,\\
(1-\tau) \cdot (\hat{q}_\tau - y), & y < \hat{q}_\tau,
\end{cases}
\label{eq:pinball}
\end{equation}
where $y$ is the ground truth and $\hat{q}_\tau$ is the predicted $\tau$-th quantile. The overall training loss is the sum of the pinball losses across all three quantile levels:
\begin{equation}
\begin{aligned}
\mathcal{L} = \frac{1}{N \cdot T_{\text{out}}} \sum_{n=1}^{N} \sum_{j=1}^{T_{\text{out}}} \Big(&\mathcal{L}_{\alpha/2}(y_{n,j}, \hat{q}_{\alpha/2}^{(n,j)}) \\
&+ \mathcal{L}_{0.5}(y_{n,j}, \hat{q}_{0.5}^{(n,j)}) \\
&+ \mathcal{L}_{1-\alpha/2}(y_{n,j}, \hat{q}_{1-\alpha/2}^{(n,j)})\Big).
\end{aligned}
\label{eq:total-loss}
\end{equation}
The median head ($\tau=0.5$) provides the point prediction $\hat{\mathbf{Y}}$, while the lower and upper heads produce quantile bounds for downstream uncertainty calibration via AS-CQR.

\subsection{Adaptive Sequential Conformalized Quantile Regression}
\label{subsec:ascqr}
Point predictions are valuable for power grid planning, but they fall short in critical grid operations that demand risk-aware decision-making. 
Existing uncertainty quantification methods face two fundamental challenges in energy systems: (1) \textbf{Heteroscedasticity}, where prediction variance correlates with consumption magnitude; and (2) \textbf{Non-stationarity}, where distribution shifts are driven by seasonality or abnormal scenarios that violate the exchangeability assumption of classical conformal prediction.
To address these challenges, we propose \textbf{Adaptive Sequential Conformalized Quantile Regression (AS-CQR)}, a distribution-free framework delivering reliable prediction intervals via two innovations:

\begin{enumerate}
    \item \textbf{Locally Adaptive Normalization}: Nonconformity scores are normalized by the predicted interval width, ensuring scale-invariant calibration.
    \item \textbf{Online Feedback Mechanism}: The target coverage level is dynamically adjusted based on recent prediction performance, enabling rapid adaptation to distribution shifts.
\end{enumerate}

\subsubsection{Locally Adaptive Nonconformity Measure}
\label{subsubsec:nonconformity}

Unlike standard Conformalized Quantile Regression (CQR)~\cite{romano2019conformalized}, which uses absolute residual errors, AS-CQR employs a normalized measure to scale the nonconformity score with the predicted interval width:
\begin{equation}
\epsilon_t = \frac{\max\left\{\hat{q}_{\text{lo}}(\mathbf{X}_t) - Y_t,\; Y_t - \hat{q}_{\text{up}}(\mathbf{X}_t)\right\}}{\hat{q}_{\text{up}}(\mathbf{X}_t) - \hat{q}_{\text{lo}}(\mathbf{X}_t) + \delta},
\label{eq:normalized_score}
\end{equation}
where $\epsilon_t \in \mathbb{R}$ is the nonconformity score at time $t$, $Y_t \in \mathbb{R}$ is the true observation, $\hat{q}_{\text{lo}}(\mathbf{X}_t)$ and $\hat{q}_{\text{up}}(\mathbf{X}_t)$ are the predicted lower and upper quantiles from Eqs.~\eqref{eq:output-lo}--\eqref{eq:output-up}, and $\delta$ is a small constant for numerical stability (e.g., $10^{-6}$).
Note that $\epsilon_t$ is a signed quantity: $\epsilon_t < 0$ when $Y_t$ lies inside the predicted interval $[\hat{q}_{\text{lo}}, \hat{q}_{\text{up}}]$, with the magnitude reflecting how deeply the observation is contained within the bounds; $\epsilon_t > 0$ when $Y_t$ falls outside, indicating miscoverage; and $\epsilon_t = 0$ when $Y_t$ lies exactly on a boundary.
Combined with the interval-width normalization, this yields a scale-invariant calibration mechanism that produces efficient prediction intervals across regions with different magnitudes as motivated in Section~\ref{sec:data}.
This normalization also makes the calibration scores more comparable across regions and time periods, preventing high-load areas from dominating the adaptation process solely because they exhibit larger absolute residuals. 

\subsubsection{Interval Construction}
\label{subsubsec:interval-construction}

At each time step $t$, the calibrated prediction interval is constructed using historical nonconformity scores. Let $\mathcal{E}_t = \{\epsilon_{t-m+1}, \ldots, \epsilon_t\}$ be a sliding window of the $m$ most recent nonconformity scores, where $m = 100$ is the window size. The correction factor is computed as:
\begin{equation}
Q_t = \text{Quantile}_{1 - \tilde{\alpha}_t}(\mathcal{E}_t),
\label{eq:quantile_correction}
\end{equation}
where $\text{Quantile}_p(\cdot)$ returns the $p$-th empirical quantile of the input set and $\tilde{\alpha}_t \in (0,1)$ is the effective miscoverage rate (detailed in Section~\ref{subsubsec:online-feedback}).
The calibrated prediction interval $\hat{C}(\mathbf{X}_t) = [\hat{C}_{\text{lo}}, \hat{C}_{\text{up}}]$ is:
\begin{equation}
\hat{C}(\mathbf{X}_t) = \left[\hat{q}_{\text{lo}}(\mathbf{X}_t) - Q_t \cdot w_t, \quad \hat{q}_{\text{up}}(\mathbf{X}_t) + Q_t \cdot w_t\right],
\label{eq:adaptive_interval}
\end{equation}
where $w_t = \hat{q}_{\text{up}}(\mathbf{X}_t) - \hat{q}_{\text{lo}}(\mathbf{X}_t) \in \mathbb{R}_{>0}$ is the raw interval width. The multiplicative correction $Q_t \cdot w_t$ ensures that interval adjustments are proportional to the model's uncertainty, maintaining the locally adaptive property.

\subsubsection{Online Feedback Calibration}
\label{subsubsec:online-feedback}

To handle distribution drift, we integrate an online update mechanism inspired by Adaptive Conformal Inference (ACI)~\cite{gibbs2021adaptive}. The key insight is to treat the target miscoverage rate $\alpha \in (0, 1)$ as a control variable that is adjusted based on recent coverage performance.
The complete online calibration procedure at each time step $t$ is:
\begin{enumerate}
    \item \textbf{Predict}: Compute quantile predictions $\hat{q}_{\text{lo}}(\mathbf{X}_t)$ and $\hat{q}_{\text{up}}(\mathbf{X}_t)$ using GE-Mamba.
    \item \textbf{Calibrate}: Construct the calibrated interval $\hat{C}(\mathbf{X}_t)$ using Eq.~\eqref{eq:adaptive_interval} with current $\tilde{\alpha}_t$.
    \item \textbf{Observe}: Receive the true observation $Y_t$.
    \item \textbf{Update}: Adjust the effective miscoverage rate for the next step.
\end{enumerate}

Let $\tilde{\alpha}_t \in (0,1)$ denote the effective miscoverage rate at time $t$, initialized as $\tilde{\alpha}_0 = \alpha$. After observing the true value $Y_t$, we update:
\begin{equation}
\tilde{\alpha}_{t+1} = \tilde{\alpha}_t + \gamma \cdot \left(\alpha - \mathbb{I}\{Y_t \notin \hat{C}(\mathbf{X}_t)\}\right),
\label{eq:aci_update}
\end{equation}
where $\gamma$ is the learning rate (0.005 in our work) that controls adaptation speed and $\mathbb{I}\{\cdot\}$ is the indicator function. This update interacts with the signed nonconformity scores to form a feedback loop:
\begin{itemize}
    \item \textbf{Undercoverage} ($Y_t \notin \hat{C}$): Since the indicator equals 1, $\tilde{\alpha}_{t+1}$ decreases by $\gamma(1-\alpha)$, raising the quantile threshold $1-\tilde{\alpha}$. Concurrently, the positive nonconformity scores from out-of-interval observations push $Q_t$ upward, jointly producing wider intervals.
    \item \textbf{Overcoverage} ($Y_t \in \hat{C}$): The indicator equals 0, so $\tilde{\alpha}_{t+1}$ increases by $\gamma\alpha$, lowering the quantile threshold. Combined with the negative nonconformity scores from well-contained observations, $Q_t$ can become negative, actively contracting the interval below the raw quantile predictions.
\end{itemize}
This feedback mechanism ensures that AS-CQR can rapidly adapt to distribution shifts for both undercoverage and overcoverage.

\subsubsection{Theoretical Guarantee}
\label{subsubsec:theory}
AS-CQR inherits the long-run coverage guarantee of adaptive conformal inference. Under mild regularity conditions, the update rule in Eq.~\eqref{eq:aci_update} ensures convergence of the average empirical coverage to the target level:
\begin{equation}
\lim_{T \to \infty} \frac{1}{T} \sum_{t=1}^{T} \mathbb{I}\{Y_t \in \hat{C}(\mathbf{X}_t)\} = 1 - \alpha.
\label{eq:coverage_guarantee}
\end{equation}
This guarantee is distribution-free, making AS-CQR particularly suitable for non-stationary energy systems where parametric assumptions may be unreliable.
\begin{table*}[ht]\small
\caption{Comparison with 15 state-of-the-art baselines on four datasets. $\downarrow$ indicates lower is better. The best results are in \textbf{bold} and the second-best are \underline{underlined}. \true~indicates target coverage ($\geq$90\%) achieved.}
\label{tab:result}
\centering
\setlength{\tabcolsep}{8pt}
\resizebox{\textwidth}{!}{%
\begin{tabular}{ll ccccc ccccc}
\toprule
& & \multicolumn{5}{c}{\textbf{Florida 1}} & \multicolumn{5}{c}{\textbf{Florida 2}} \\
\cmidrule(lr){3-7} \cmidrule(lr){8-12}
\textbf{Category} & \textbf{Method} & MAE$\downarrow$ & RMSE$\downarrow$ & MPIW$\downarrow$ & IS$\downarrow$ & COV & MAE$\downarrow$ & RMSE$\downarrow$ & MPIW$\downarrow$ & IS$\downarrow$ & COV \\
\midrule
\multirow{6}{*}{GNN} & DCRNN & 47.24 & 78.39 & 181.93 & 312.35 & \false & 44.20 & 73.87 & 170.24 & 295.39 & \false \\
 & STGCN & 48.56 & 80.49 & \underline{130.17} & 351.78 & \false & 43.40 & 70.29 & 145.76 & 274.06 & \false \\
 & AGCRN & 60.99 & 103.49 & 225.29 & 384.37 & \false & 37.97 & 61.76 & 127.92 & 258.03 & \false \\
 & STZINB & 79.92 & 134.67 & 340.00 & 460.18 & \false & 98.84 & 163.50 & 348.65 & 606.27 & \false \\
 & UQGNN & 47.37 & 78.09 & 204.44 & 281.15 & \true & 39.17 & 63.50 & 189.37 & 245.14 & \true \\
 & TrustEnergy & 59.13 & 93.48 & 235.40 & 369.17 & \false & 56.02 & 86.93 & 219.78 & 333.47 & \false \\
\midrule
\multirow{2}{*}{Attention} & DSTAGNN & 126.09 & 180.56 & 163.58 & 1999.85 & \false & 116.91 & 167.53 & 178.11 & 1841.64 & \false \\
 & ASTGCN & 50.52 & 83.76 & 197.97 & 334.10 & \false & 43.18 & 72.02 & 172.14 & 297.13 & \false \\
\midrule
\multirow{3}{*}{Transformer} & GluonTS & 152.45 & 241.90 & 184.42 & 1396.62 & \false & 161.55 & 224.24 & 134.22 & 2082.87 & \false \\
 & PatchTST & 386.23 & 532.86 & 189.64 & 6789.47 & \false & 53.84 & 83.91 & 262.86 & 365.57 & \false \\
 & PowerPM & 50.91 & 83.98 & 197.68 & 314.82 & \false & 41.83 & 67.39 & 170.23 & 267.08 & \false \\
\midrule
\multirow{2}{*}{LLM} & ST-LLM & \underline{38.49} & \underline{64.27} & 134.28 & \underline{246.66} & \true & \underline{34.12} & \underline{55.42} & \underline{114.71} & \underline{210.64} & \true \\
 & UrbanGPT & 44.56 & 72.98 & 164.60 & 293.87 & \false & 36.81 & 57.42 & 140.58 & 243.29 & \false \\
\midrule
\multirow{2}{*}{Mamba} & G-Mamba & 43.13 & 71.22 & 152.66 & 295.48 & \true & 37.03 & 60.47 & 143.06 & 245.46 & \false \\
 & U-Mamba & 42.55 & 73.76 & 134.05 & 312.58 & \false & 34.58 & 55.86 & 115.93 & 221.87 & \true \\
\midrule
Ours & \textbf{EnergyMamba} & \textbf{36.57} & \textbf{61.06} & \textbf{122.51} & \textbf{231.86} & \true & \textbf{32.42} & \textbf{52.66} & \textbf{107.83} & \textbf{198.23} & \true \\
\midrule\midrule
& & \multicolumn{5}{c}{\textbf{New York}} & \multicolumn{5}{c}{\textbf{California}} \\
\cmidrule(lr){3-7} \cmidrule(lr){8-12}
\textbf{Category} & \textbf{Method} & MAE$\downarrow$ & RMSE$\downarrow$ & MPIW$\downarrow$ & IS$\downarrow$ & COV & MAE$\downarrow$ & RMSE$\downarrow$ & MPIW$\downarrow$ & IS$\downarrow$ & COV \\
\midrule
\multirow{6}{*}{GNN} & DCRNN & 52.03 & 80.09 & 255.49 & 364.27 & \false & 370.53 & 638.21 & 777.09 & 2634.64 & \false \\
 & STGCN & 70.02 & 107.28 & 224.87 & 473.28 & \false & 285.81 & 551.64 & 1266.54 & 1651.27 & \false \\
 & AGCRN & 51.75 & 79.31 & 221.29 & 352.53 & \false & 272.77 & 507.80 & 1173.14 & 1537.77 & \false \\
 & STZINB & 67.44 & 107.20 & 315.22 & 421.86 & \false & 430.81 & 735.41 & 2499.45 & 2659.50 & \false \\
 & UQGNN & 64.71 & 95.55 & 360.78 & 439.40 & \false & 306.43 & 551.83 & 2036.20 & 2108.05 & \false \\
 & TrustEnergy & 124.04 & 178.58 & 742.06 & 930.46 & \false & 477.95 & 787.52 & 1875.89 & 3997.81 & \false \\
\midrule
\multirow{2}{*}{Attention} & DSTAGNN & 63.14 & 99.22 & 380.22 & 439.31 & \false & 1493.13 & 2150.30 & 1402.32 & 2697.25 & \false \\
 & ASTGCN & 52.84 & 81.62 & 253.81 & 378.99 & \false & 257.08 & 480.29 & 1151.04 & 1500.21 & \true \\
\midrule
\multirow{3}{*}{Transformer} & GluonTS & 62.87 & 105.16 & 446.82 & 467.94 & \false & 2840.59 & 4284.23 & 1469.21 & 6346.11 & \false \\
 & PatchTST & 103.94 & 157.12 & 781.38 & 783.94 & \false & 3848.31 & 5490.15 & 363.90 & 7094.35 & \false \\
 & PowerPM & 51.52 & 76.82 & 250.03 & 343.51 & \true & 222.01 & 439.27 & 1161.55 & 1439.98 & \true \\
\midrule
\multirow{2}{*}{LLM} & ST-LLM & 51.25 & 78.41 & 211.39 & 335.99 & \false & 265.94 & 519.28 & 761.72 & 1959.38 & \false \\
 & UrbanGPT & 46.42 & 69.94 & 215.50 & 332.61 & \true & 210.55 & \underline{423.03} & 740.19 & 1563.31 & \true \\
\midrule
\multirow{2}{*}{Mamba} & G-Mamba & \underline{45.40} & \underline{68.20} & \textbf{190.12} & 318.37 & \true & \underline{209.66} & 437.21 & 1034.89 & \textbf{1265.76} & \true \\
 & U-Mamba & 46.82 & 71.69 & 220.56 & \underline{313.85} & \true & 240.71 & 428.54 & \underline{735.47} & 1443.22 & \true \\
\midrule
Ours & \textbf{EnergyMamba} & \textbf{43.13} & \textbf{64.79} & \underline{201.87} & \textbf{295.42} & \true & \textbf{199.18} & \textbf{401.88} & \textbf{692.05} & \underline{1279.17} & \true \\
\bottomrule
\end{tabular}
}
\end{table*}

\section{Evaluation}\label{evaluation}

In this section, we conduct a comprehensive experimental evaluation of our proposed EnergyMamba. Specifically, we aim to address the following five research questions:
\begin{itemize}
\item \textbf{RQ 1}: How does EnergyMamba perform compared to baselines?
\item \textbf{RQ 2}: Is EnergyMamba effective for uncertainty quantification?
\item \textbf{RQ 3}: Are all components in EnergyMamba effective?
\item \textbf{RQ 4}: How does EnergyMamba perform under abnormal scenarios such as extreme weather events?
\item \textbf{RQ 5}: Is EnergyMamba computationally efficient?
\end{itemize}

\subsection{Evaluation Setup}\label{setup}

\subsubsection{Datasets}
We evaluate the performance of EnergyMamba on four real-world energy consumption datasets. 
The first two datasets originate from Florida, covering 201 census block groups (CBGs) with fine-grained data collected at 30-minute intervals. 
We denote the dataset spanning the year 2018 as Florida 1 and the dataset spanning 2019 as Florida 2. 
To further verify the generalizability, we incorporate two additional datasets from New York \cite{nyiso} and California \cite{caiso} in 2024, both recorded at 1-hour intervals in a coarser spatial granularity (11 regions and 9 zones, respectively).
More details will be in Appendix \ref{sec:app_data}.

\subsubsection{Baselines} 
We compare EnergyMamba with 5 categories of 15 state-of-the-art baselines: 
(1) GNN-based: DCRNN \cite{dcrnn}, STGCN \cite{stgcn}, AGCRN \cite{bai2020adaptive}, STZINB \cite{stzinb-gnn}, UQGNN \cite{yu2025uqgnn}, TrustEnergy \cite{yu2026trustenergy}; 
(2) Attention-based: DSTAGNN \cite{lan2022dstagnn}, ASTGCN \cite{guo2019attention}; 
(3) Transformer-based: GluonTS \cite{alexandrov2020gluonts}, PatchTST \cite{nie2022time}, PowerPM \cite{tu2024powerpm}; 
(4) LLM-based: ST-LLM \cite{liu2024st}, UrbanGPT \cite{li2024urbangpt};
(5) Mamba-based: G-Mamba \cite{Graph-Mamba}, U-Mamba \cite{ma2024u}. Details are shown in Appendix \ref{app_baseline}.

\subsubsection{Metrics} 
We utilize Mean Absolute Error (MAE) and Root Mean Squared Error (RMSE) to evaluate the performance of deterministic prediction, and three other commonly used metrics (i.e., Mean Prediction Interval Width (MPIW), Interval Score (IS), and Coverage (COV)), to evaluate the performance of uncertainty quantification. Details in Appendix~\ref{app_metrics}. 

\subsubsection{Implementation Details}
All experiments are conducted on a Linux server with an NVIDIA A100 GPU (80 GB). We use Adam with batch size 128 and initial learning rate \(1\times10^{-3}\), decayed every 15 epochs, and apply early stopping with patience 50 based on validation loss. Datasets are split into training/validation/testing sets at 8:1:1, with input and prediction lengths of 192 and 6 time steps, corresponding to 4 days/3 hours for Florida and 8 days/6 hours for NYISO and CAISO.
For EnergyMamba, we set hidden dimension \(D=64\), Mamba expansion factor 2, SSM state dimension \(D_s=16\), and \(S=2\) encoder-decoder stages with \(K=2\) GE-Mamba blocks per stage. The model is trained end-to-end with Eq.~\eqref{eq:total-loss} using \(\tau\in\{0.05,0.5,0.95\}\); for AS-CQR, we set \(\alpha=0.1\), \(m=100\), \(\gamma=0.005\), dropout 0.1, and gradient clipping 1.0.
For deterministic baselines (e.g., DCRNN, STGCN, and AGCRN), we replace their original output layers with the same quantile heads used in EnergyMamba and train them with the same composite pinball loss for fair comparison.

\subsection{Overall Performance Comparison (RQ 1)}
An overall comparison of EnergyMamba and other baselines is presented in Table \ref{tab:result}. We found that our EnergyMamba consistently achieves the best performance across most metrics for both prediction accuracy and uncertainty quantification. Specifically, our framework reduces MAE by approximately 5\% compared to the best baseline based on the average of all four datasets.
In addition, EnergyMamba also demonstrates superior performance on uncertainty quantification, with around 6\% improvement in IS and reaching the target coverage.

\begin{figure}[ht]
\centering
\subfigure[Selective regression.]{\label{curve1}
\includegraphics[width=0.48\linewidth]{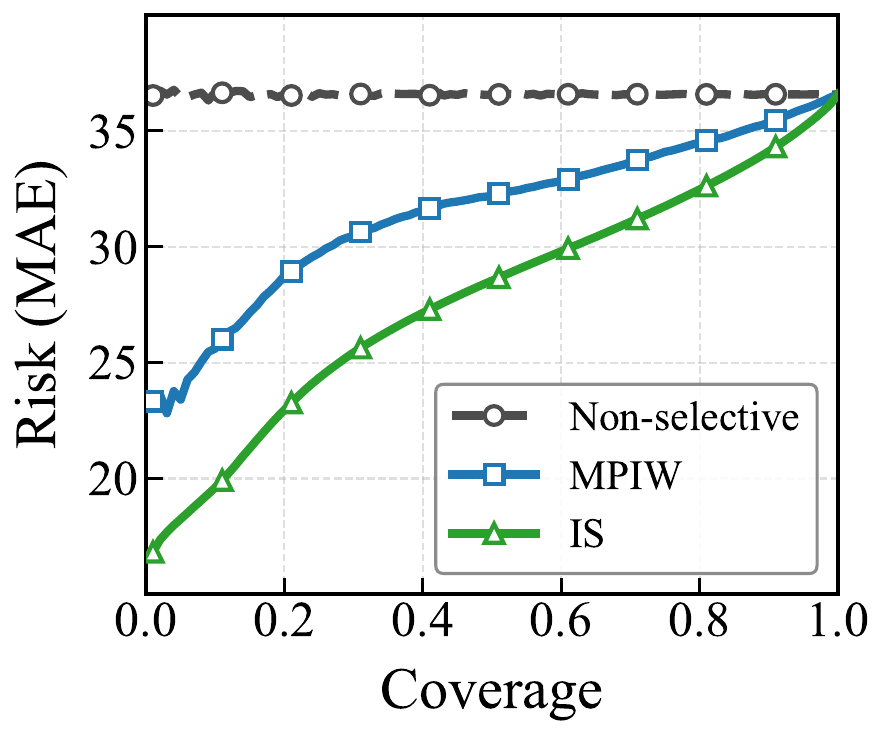}}
\subfigure[Ideal vs. Empirical Coverage.]{\label{curve2}
\includegraphics[width=0.48\linewidth]{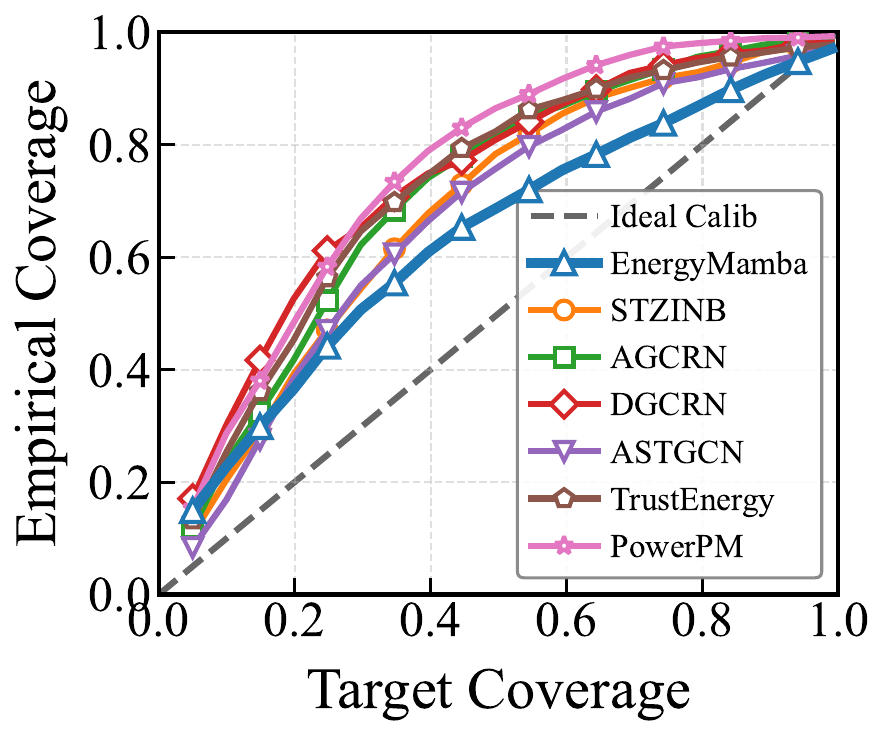}}
\caption{UQ analyses on the Florida dataset 1.}
\Description{Two uncertainty-quantification plots for Florida 1: a selective regression curve and an ideal-versus-empirical coverage comparison.}
\label{curve_figure}
\end{figure}

\subsection{Effectiveness of UQ (RQ 2)}
Furthermore, we adopt selective regression~\cite{sokol2024conformalized,shah2022selective} to evaluate uncertainty quantification performance. Selective regression allows the model to abstain from making predictions when confidence is insufficient, and is characterized by \emph{coverage} (the fraction of samples selected for prediction, distinct from COV, which denotes the proportion of ground truth observations falling within the predicted bounds) and \emph{risk} (measured by MAE).
As shown in Figure~\ref{curve1}, in the absence of uncertainty quantification, the MAE remains nearly constant across different coverage levels. In contrast, when uncertainty is taken into account, the error increases as coverage grows, indicating a positive correlation between prediction error and coverage.
Moreover, Figure~\ref{curve2} shows that EnergyMamba achieves the most reliable calibration, as its curve lies closest to the diagonal line. These results confirm that the proposed uncertainty estimates are informative and meaningful.

\subsection{Ablation Study (RQ 3)}
To evaluate component contributions, we compare EnergyMamba against four variants: (1) \textbf{w/o GCN}: Removes spatial context, treating nodes independently; (2) \textbf{w/o BIP}: Uses unidirectional processing, ignoring backward dependencies; (3) \textbf{w/o U-Net}: Uses a flat stack, removing multi-scale modeling; (4) \textbf{w/o AS-CQR}: Replaces adaptive mechanisms with standard CQR; and (5) \textbf{GCN + Linear (decoupled)}: Applying GCN only at the input and output layers.
Table \ref{tab:ablation} confirms the effectiveness of each component. Removing GCN significantly degrades performance, validating the necessity of spatial context. The higher MAE in w/o BIP and w/o U-Net highlights the importance of backward dependencies and hierarchical temporal modeling, respectively. Also, w/o AS-CQR yields higher MPIW and misses target coverage, proving that adaptive normalization is critical for reliable uncertainty quantification.
Finally, the decoupled design fails to capture evolving spatial dependencies, leading to degradation in performance.

\subsection{Case Studies for Abnormal Situations (RQ 4)}
\begin{figure}[tb]
\centering
\subfigure[Hurricane.]{
\includegraphics[width=0.48\linewidth]{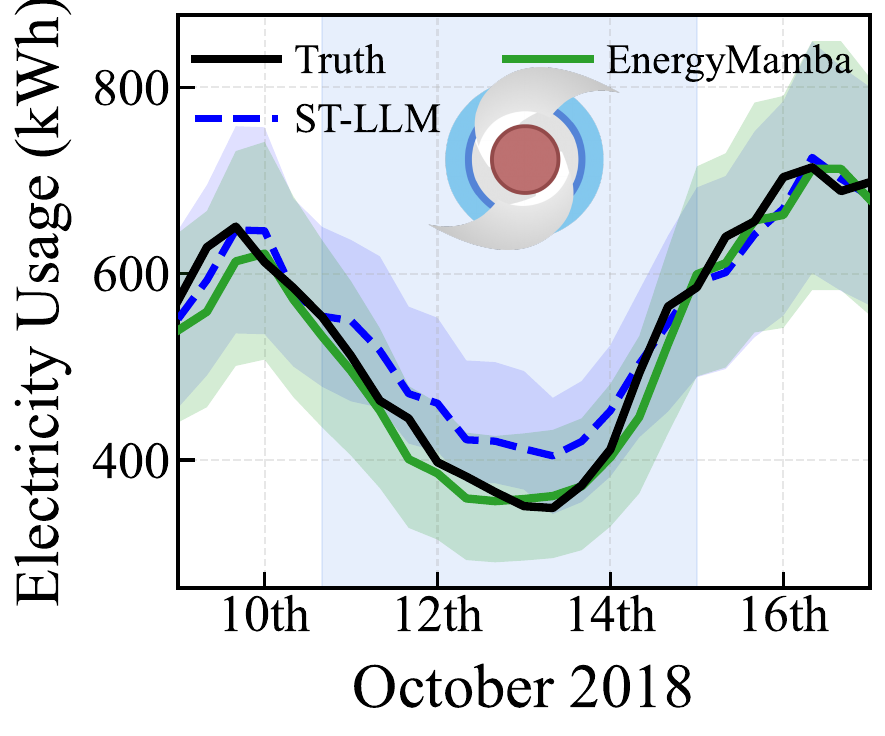}}
\subfigure[Heat Wave.]{
\includegraphics[width=0.48\linewidth]{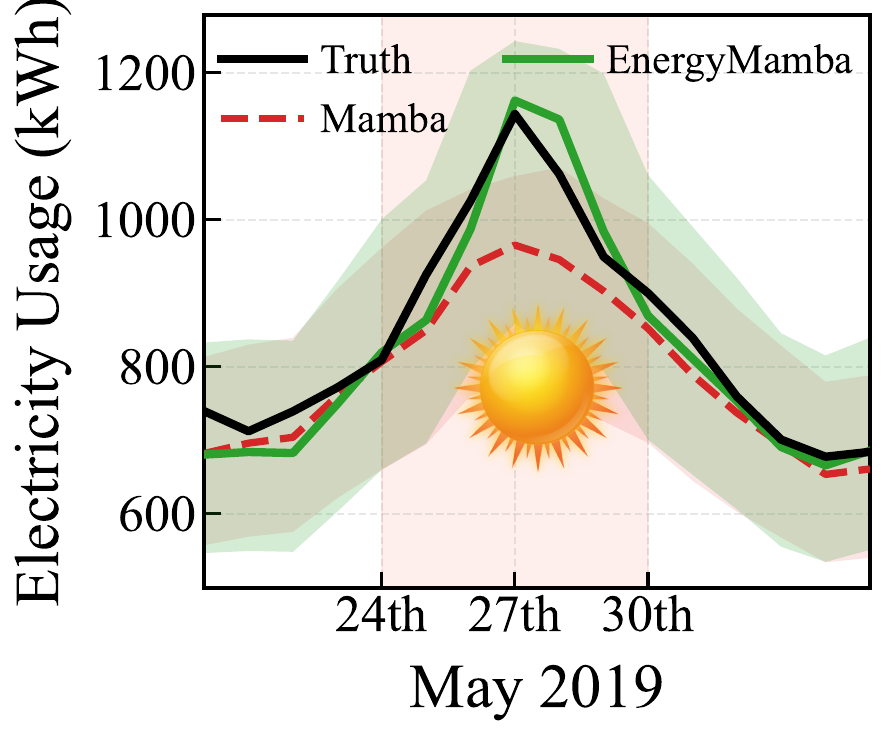}}
\caption{Prediction results under extreme events.}
\Description{Two case-study plots show prediction intervals under a hurricane and a heat wave, comparing predicted trajectories with observed values during abnormal events.}
\label{anomaly}
\end{figure}

One of the most challenging aspects of energy consumption prediction is handling abnormal or extreme scenarios, such as hurricanes, heat waves, and cold snaps, that often lead to sudden and significant deviations in consumption patterns. 
Accurate and reliable prediction under these extreme events is critical for ensuring grid stability, optimizing energy resource allocation, and enabling timely decision-making for both utilities and emergency response teams.
As shown in Figure \ref{anomaly}, under different extreme events, EnergyMamba successfully maintained accurate predictions, and ground truth values are within the prediction intervals in most cases, whereas the best baseline struggled to adapt to these abrupt changes. These results highlight EnergyMamba's superior adaptability to both demand drops and surges under abnormal scenarios.

\begin{table}[ht]\small
\caption{Ablation study on the Florida dataset 1.}
\label{tab:ablation}
\centering
\setlength{\tabcolsep}{6pt}
\begin{tabular}{@{}lccccc@{}}
\toprule
\textbf{Variants} & MAE$\downarrow$ & RMSE$\downarrow$& MPIW$\downarrow$ & IS$\downarrow$ & COV \\ 
\midrule
w/o GCN & 47.82 & 76.14 & 145.27 & 298.93 & \false \\ 
w/o BIP & 42.91 & 74.52 & 139.68 & 257.15 & \false \\
w/o U-Net & 39.24 & 65.87 & 141.45 & 271.72 & \false \\
w/o AS-CQR & 44.12 & 72.08 & 152.36 & 263.47 & \false \\
GCN + Linear & 58.79 & 105.23 & 220.95 & 385.71 & \false \\
\midrule
\textbf{EnergyMamba} & \textbf{36.57} & \textbf{61.06} & \textbf{122.51} & \textbf{231.86} & \true \\ 
\bottomrule
\end{tabular}
\end{table}

\begin{table}[ht]\small
\caption{Computational complexity comparison.}
\label{tab:complexity}
\centering
\setlength{\tabcolsep}{2pt}
\resizebox{0.48\textwidth}{!}{%
\begin{tabular}{@{}ll cccc@{}}
\toprule
\textbf{Category} & \textbf{Method} & Train (s)$\downarrow$ & Infer (s)$\downarrow$ & Mem (MB)$\downarrow$ & Params (K)$\downarrow$ \\
\midrule
\multirow{6}{*}{GNN} 
 & DCRNN    & 15847 & 2.75 & 1108 & 23.7 \\
 & STGCN    & 15570 & 2.66 & 6308 & 438.1 \\
 & AGCRN    & 662 & 1.25 & 3781 & 747.4 \\
 & STZINB   & 1913 & 3.80 & 8109 & 249.9 \\
 & UQGNN    & 710 & 3.10 & 4039 & 297.7 \\
 & TrustEnergy    & 3865 & 3.90 & 27119 & 1012.2 \\
\midrule
\multirow{2}{*}{Attention} 
 & DSTAGNN  & 3291 & 2.20 & 2190 & 298.6 \\
 & ASTGCN   & 1759 & 2.15 & 2125 & 230.1 \\
\midrule
\multirow{3}{*}{Transformer} 
 & GluonTS  & 104 & 0.95 & 1378 & 25.9 \\
 & PatchTST & 152 & 1.85 & 3167 & 145.4 \\
 & PowerPM & 8900 & 2.60 & 16500 & 11085.4 \\
\midrule
\multirow{2}{*}{LLM} 
 & ST-LLM   & 13520 & 5.51 & 24850 & 67462.5 \\
 & UrbanGPT & 14200 & 6.00 & 28600 & 72520.8 \\
\midrule
\multirow{2}{*}{Mamba} 
 & G-Mamba    & 6285 & 2.89 & 7499 & 299.5 \\
 & U-Mamba  & 11090 & 3.60 & 12582 & 664.3 \\
\midrule
Ours & \textbf{EnergyMamba} & 7945 & 2.78 & 4125 & 312.8 \\
\bottomrule
\end{tabular}
}
\end{table}

\subsection{Computational Complexity Analysis (RQ 5)}
We explicitly evaluate the computational efficiency of EnergyMamba in terms of training time, peak GPU memory usage, and model size. Table~\ref{tab:complexity} reports detailed comparisons with representative baselines.

\textbf{Time Complexity}
EnergyMamba completes training in 7,945 seconds. This efficiency primarily comes from the Mamba backbone, which scales linearly with sequence length ($O(T)$), avoiding the quadratic cost of attention-based alternatives.
Compared with LLM-based baselines, EnergyMamba is substantially faster and thus more practical for frequent retraining in dynamic grid scenarios. 
Furthermore, during the inference phase, EnergyMamba achieves a highly competitive latency. It operates at approximately twice the speed of heavy LLM models and also outperforms other state-of-the-art Mamba variants such as G-Mamba and U-Mamba. 

\textbf{Memory Footprint.}
The peak GPU memory usage of EnergyMamba is 4.03 GB, which is notably lower than many attention-based and LLM-based baselines. This moderate memory demand improves deployability in resource-constrained operational environments.

\textbf{Parameter Efficiency.}
With 312.8K trainable parameters, EnergyMamba remains lightweight while still achieving superior predictive performance (RQ 1), indicating that gains come from architectural design rather than brute-force model scaling.
Overall, EnergyMamba achieves a favorable balance between accuracy and efficiency.

\section{Related Work}\label{literature review}

\subsection{Energy Consumption Prediction}
Energy consumption prediction has attracted considerable attention from both academia and industry. 
Early energy consumption prediction approaches largely relied on statistical methods, such as linear regression~\cite{hong2011naive}, support vector regression~\cite{sapankevych2009time}, and random forest regression~\cite{wu2015power}, which model linear relationships between energy consumption.
However, these methods often struggle to capture the complex, nonlinear dynamics inherent in energy consumption data.
To address this limitation, deep learning approaches have been widely adopted, including general neural forecasting methods~\cite{shen2026credit, cheng2025bts, yu2025uqgnn, yu2026health, shen2026cited}, Transformers~\cite{alexandrov2020gluonts, shen2025learning, buratto2024seq2seq}, and Diffusion~\cite{xu2026geogen,xu2025autostdiff,xu2026synhat}, which excel at mining temporal patterns from historical time series.
More recently, the emergence of Foundation Models~\cite{tu2024powerpm} and Large Language Models~\cite{li2026llmclinicalgraphstructure, 10.1145/3748636.3763223, liang2025energygpt} has introduced a new paradigm, leveraging extensive pre-training to achieve superior generalization.
Despite these advancements, most existing works formulate energy consumption prediction as a purely time-series prediction problem, neglecting the intrinsic spatial dependencies among different regions or grid zones that could provide critical information. Furthermore, the reliability of predictions, which is crucial for real-world decision-making under extreme weather or grid anomalies, remains underexplored.

\subsection{Uncertainty-aware Spatiotemporal Prediction}
Quantifying uncertainty is significant for robust and reliable decision-making~\cite{yang2023re,yang2024regulating}. 
Most existing uncertainty-aware spatiotemporal prediction works~\cite{li2026fast,xiao2025multifrequency} focus on distribution-based methods or conformal prediction.
Distribution-based methods assume that the target variable follows a specific probability distribution and optimize the model to predict the distribution parameters. For example, 
UQGNN~\cite{yu2025uqgnn} assumes a Gaussian distribution, while STZINB~\cite{stzinb-gnn} leverages a zero-inflated negative binomial distribution.
Conformal Prediction is a distribution-free framework that constructs valid prediction intervals. CF-GNN~\cite{huang2024uncertainty} integrates conformal prediction with GNN, while TrustEnergy~\cite{yu2026trustenergy} further enhances this by incorporating meta-learning to achieve context-aware prediction.
Despite these advances, a critical gap remains: most existing methods assume data exchangeability or stationarity, which do not hold in real-world energy systems subject to distribution shifts.
They typically lack mechanisms to dynamically adjust uncertainty bounds in response to online feedback, leading to potential miscoverage during abnormal or extreme events. 
Our work addresses this by incorporating adaptive conformal inference into a spatiotemporal framework, ensuring robust calibration even under non-stationary conditions.

\section{Conclusion}\label{conclusion}
In this paper, we propose EnergyMamba, an uncertainty-aware spatiotemporal framework for accurate and reliable energy consumption prediction. The design of EnergyMamba is directly motivated by empirical insights, addressing key challenges including physical spatial dependencies and load-dependent heteroscedasticity. EnergyMamba consists of two core components: (i) GE-Mamba, which integrates GCN with a Selective State Space Model within a U-Net architecture. By leveraging a Bidirectional Processing mechanism and multi-scale feature extraction, GE-Mamba effectively captures complex spatiotemporal patterns while maintaining linear computational complexity; and (ii) AS-CQR, which includes locally adaptive normalization and online feedback mechanisms to handle non-stationarity, providing reliable uncertainty estimates even under distribution shifts during extreme weather events.
Extensive experiments on four real-world datasets from Florida, New York, and California demonstrate that EnergyMamba outperforms 15 state-of-the-art baselines, improving prediction accuracy by around 5\% and uncertainty quantification by 6\%.

\section{Limitations and Ethical Considerations}
The current implementation of EnergyMamba operates at the regional level (e.g., CBGs) and does not explicitly model the physical topology of the power grid. While this design is easily generalizable across different regions and avoids the need to track grid topology changes, it may overlook structural and operational dependencies. As future work, this framework can be extended to incorporate grid topology and leverage adaptive graph learning to dynamically update network representations.

We adhere to the KDD Code of Ethics. De-identified data was obtained from a municipal utility provider in Florida under an NDA, which is stored and processed exclusively on FSU's secured computing facilities. We declare no conflicts of interest or foreseeable risks.

\section*{Acknowledgment}
We thank all the reviewers for their insightful feedback to improve this paper.
This work is partially supported by the FSU Startup Fund, FSU CRC Summer Research Support (SRS) Award Program, and
FSU/AWS Research Acceleration Fund.

\clearpage
\newpage
\bibliographystyle{ACM-Reference-Format}
\balance
\bibliography{reference}

\appendix
\section*{Appendix}

\section{Data-driven Analysis}\label{sec:app_data}
In this part, we provide detailed descriptions of data collection, preprocessing, and analysis procedures.
The four datasets span different spatial regions and scales, temporal resolutions, and time periods, enabling a comprehensive evaluation of our framework's generalizability. A summary of datasets is provided in Table~\ref{table datasets}.

\subsection{Florida Datasets (Florida 1 \& Florida 2)}
We have access to utility data from a municipal utility provider in Florida. The data include household-level electricity consumption from smart meters. These meters automatically record electricity consumption at 30-minute intervals and transmit the readings to a central data management system. The raw data includes meter ID, timestamp, and energy consumption in kilowatt-hours (kWh). 
We partition the Florida data into two distinct datasets in order to verify the model's generalization capabilities under different distribution shifts, e.g., different types of extreme weather events (Hurricane Michael in 2018 and record-breaking heatwaves in 2019). Florida 1 spans the entire year of 2018 (January 1 to December 31), while Florida 2 covers the year 2019. Each dataset contains approximately 17,520 time steps (48 readings per day $\times$ 365 days).
We extensively evaluated our framework on a full 10-year continuous dataset, confirming its stability under regular multi-year seasonal cycles. Since standard seasonality is relatively predictable, we deliberately selected 2018 and 2019 for the main manuscript to rigorously stress-test the model against extreme distribution shifts. These years exhibit radically different weather conditions: Florida 1 includes Hurricane Michael (October 2018), while Florida 2 features severe heat waves (May 2019).

At the utility scale, our 5\% system-wide accuracy improvement can translate to massive cost savings by reducing reliance on spinning reserves and mitigating millions in over-procurement penalties. Furthermore, EnergyMamba achieves an \textasciitilde6\% improvement in uncertainty quantification (Interval Score). Because modern grid operations are deeply risk-sensitive, this enhanced probabilistic reliability is critical for determining safe reserve margins and preventing blackouts during extreme events. By jointly optimizing deterministic accuracy and probabilistic forecasting, EnergyMamba delivers a dual contribution with profound economic and operational value.



\begin{table}[ht]\small
\caption{Summary of four real-world energy consumption datasets used in this study. } 
\label{table datasets}
\centering
\setlength{\tabcolsep}{4pt}
\begin{tabular}{c|cccccc}
\toprule
Dataset   & Year  & Nodes & Resolution & Time Steps & Unit\\ 
\hline
Florida 1 & 2018 & 201 CBGs & 30 min & 17,520 & kWh \\
Florida 2 & 2019 & 201 CBGs & 30 min & 17,520 & kWh \\
NYISO &   2024 & 11 Zones & 1 hour & 8,760 & MWh\\
CAISO &   2024 & 9 Regions & 1 hour & 8,760 & MWh\\
\bottomrule
\end{tabular}
\end{table}

\subsection{New York ISO Dataset (NYISO)}
The New York dataset is sourced from the New York Independent System Operator (NYISO). 
NYISO is collected from supervisory control and data acquisition (SCADA) systems that monitor real-time power flows across the transmission network. Load readings are aggregated at the zonal level, where each zone represents a distinct load pocket with similar electrical characteristics.
We collected data for the year 2024, recorded at 1-hour intervals, resulting in 8,760 time steps per zone. The hourly resolution aligns with standard electricity market operations and enables evaluation on coarser-grained temporal patterns.

\subsection{California ISO Dataset (CAISO)}
The California dataset is obtained from the California Independent System Operator (CAISO), which oversees the operation of California's bulk electric power system.
The dataset covers 9 Transmission Access Charge (TAC) areas: Pacific Gas \& Electric (PG\&E), Southern California Edison (SCE), San Diego Gas \& Electric (SDG\&E), Valley Electric Association (VEA), and several smaller municipal utilities. These regions represent California's three major investor-owned utilities and associated service territories.
Similar to the NYISO dataset, we collected hourly load data for the year 2024, yielding 8,760 time steps per region.

\subsection{Data Preprocessing and Management}
We applied consistent preprocessing procedures across all datasets to ensure data quality and compatibility with our modeling framework.

\textbf{Missing Value Handling.} For the Florida datasets, missing readings (0.3\% of data) were imputed using linear interpolation between adjacent time steps. For NYISO and CAISO datasets, missing values were rare ($<$0.1\%) and were filled using the same imputation strategy.

\textbf{Outlier Detection.} We identified outliers using the interquartile range (IQR) method. Values exceeding 1.5 times the IQR beyond the first or third quartile were flagged and manually inspected. Genuine extreme values (e.g., during heat waves) were retained, while obvious measurement errors were corrected using temporal interpolation.

\textbf{Normalization.} To ensure stable training, we applied log normalization to scale consumption values for each node independently. All time series data are normalized using a transformation of natural logarithm, which is represented as:
\begin{equation}
    X'=\ln(X+1),
\end{equation}
where $X$ denotes the original dataset and $X'$ is the normalized dataset. All values are incremented by 1 to avoid undefined logarithms for zero-valued entries.

\textbf{Graph Construction.} Following Section~\ref{subsec:graph-construction}, we constructed adjacency matrices based on geographical distances between region centroids as a proximity-based coupling proxy. For Florida datasets, we used CBG centroids; for NYISO and CAISO, we used the geographical centers of each zone/region.

\section{Experiment Setup}\label{app_experiment}

\subsection{Baseline}\label{app_baseline}
These baselines cover graph, attention, Transformer, LLM, and state-space paradigms, and include both deterministic and probabilistic forecasting methods (STZINB, UQGNN, and TrustEnergy).

\paragraph{GNN-based methods.}
\textbf{DCRNN}~\cite{dcrnn} is a diffusion-convolution seq2seq model; \textbf{STGCN}~\cite{stgcn} combines graph and temporal convolutions; \textbf{AGCRN}~\cite{bai2020adaptive} learns adaptive node-wise dependencies; \textbf{STZINB}~\cite{stzinb-gnn} is a probabilistic graph model with a zero-inflated negative binomial formulation; \textbf{UQGNN}~\cite{yu2025uqgnn} focuses on multivariate uncertainty quantification; and \textbf{TrustEnergy}~\cite{yu2026trustenergy} combines meta-learning with conformal prediction. 

\paragraph{Attention-based methods.}
\textbf{DSTAGNN}~\cite{lan2022dstagnn} is an attention-based dynamic graph model, while \textbf{ASTGCN}~\cite{guo2019attention} introduces explicit spatial-temporal attention into graph convolution. 

\paragraph{Transformer-based methods.}
\textbf{GluonTS}~\cite{alexandrov2020gluonts} provides strong probabilistic forecasting baselines, \textbf{PatchTST}~\cite{nie2022time} is a patch-based Transformer for long-range multivariate forecasting, and \textbf{PowerPM}~\cite{tu2024powerpm} is a pre-trained power forecasting model based on masked modeling and contrastive learning. 

\paragraph{LLM-based methods.}
\textbf{ST-LLM}~\cite{liu2024st} is a spatial-temporal large language model for traffic forecasting, and \textbf{UrbanGPT}~\cite{li2024urbangpt} is an instruction-tuned urban forecasting framework built on an LLM backbone. 

\paragraph{Mamba-based methods.}
\textbf{G-Mamba}~\cite{Graph-Mamba} combines graph modeling with Mamba blocks, while \textbf{U-Mamba}~\cite{ma2024u} adopts a U-shaped Mamba architecture for hierarchical sequence modeling. This category is the closest architectural family to EnergyMamba and is therefore especially relevant for isolating the contribution of our graph-enhanced design.

\subsection{Metrics}\label{app_metrics}
We report MAE and RMSE for deterministic prediction, and Mean Prediction Interval Width (MPIW), Interval Score (IS), and Coverage (COV) for uncertainty quantification. Here, $y_i$ and $\hat{y}_i$ denote the ground truth and prediction for the $i$-th sample, and $[l_i,u_i]$ denotes its prediction interval. MAE and RMSE measure point-forecast accuracy, with RMSE assigning a larger penalty to large deviations and thus being more sensitive to peak-load errors. For interval prediction, MPIW measures sharpness, \(\text{MPIW}=\frac{1}{N}\sum_{i=1}^{N}(u_i-l_i)\), Coverage measures calibration, and IS summarizes both by penalizing intervals that are either too wide or fail to cover the ground truth. Reporting these metrics together avoids favoring trivially wide intervals that attain high coverage but poor usefulness.
\begin{equation}
\text{MAE}=\frac{1}{N}\sum_{i=1}^{N}|y_i-\hat{y}_i|, \quad
\text{RMSE}=\sqrt{\frac{1}{N}\sum_{i=1}^{N}(y_i-\hat{y}_i)^2}.
\end{equation}
\begin{equation}
\text{IS}=\frac{1}{N}\sum_{i=1}^{N}\Big[(u_i-l_i)
+ \frac{2}{\alpha}(l_i-y_i)\mathbb{I}(y_i<l_i)
+ \frac{2}{\alpha}(y_i-u_i)\mathbb{I}(y_i>u_i)\Big].
\end{equation}
\begin{equation}
\text{Coverage}=\frac{100\%}{N}\sum_{i=1}^{N}\mathbb{I}(l_i\le y_i\le u_i).
\end{equation}
Lower MAE, RMSE, MPIW, and IS are better, while Coverage should be close to the target confidence level. In practice, a good uncertainty estimator should achieve reliable Coverage without excessively increasing MPIW. If Coverage is substantially below the target level, the intervals are under-dispersed and the model is overconfident; if it is much higher than the target, the intervals are often overly conservative.

\section*{GenAI Disclosure}
In the preparation of this work, the authors utilized Generative AI tools solely for the purpose of language refinement and improving readability. No AI tools were used to generate scientific concepts, experimental results, or the intellectual content of this paper. The authors have reviewed all AI-assisted edits and take full responsibility for the final content of the manuscript.

\end{document}